\documentclass{article}

\usepackage{arxiv}

\usepackage[utf8]{inputenc} % allow utf-8 input
\usepackage[T1]{fontenc}    % use 8-bit T1 fonts
\usepackage{hyperref}       % hyperlinks
\usepackage{url}            % simple URL typesetting
\usepackage{booktabs}       % professional-quality tables
\usepackage{amsfonts}       % blackboard math symbols
\usepackage{nicefrac}       % compact symbols for 1/2, etc.
\usepackage{microtype}      % microtypography
\usepackage{lipsum}
\usepackage{graphicx}
\usepackage{amssymb, amsmath, amsthm}
\usepackage[caption=false]{subfig}
\usepackage{tikz}
\usepackage{multirow}
\usetikzlibrary{shadows,arrows}

\newtheorem{definition}{Definition}

\title{Robust reconstructions by multi-scale/irregular tangential covering}

\author{
  Antoine Vacavant\thanks{http://antoine-vacavant.eu} \\
  Universit\'e Clermont Auvergne, CNRS, SIGMA Clermont\\ Institut Pascal \\ F-63000 Clermont-Ferrand, France \\
  \texttt{antoine.vacavant@uca.fr} \\
   \And
 Bertrand Kerautret \\
  Univ Lyon, Lyon 2\\
  LIRIS\\
  F-69676 Lyon, France \\
  \texttt{bertrand.kerautret@univ-lyon2.fr} \\ 
  \And 
  Fabien Feschet \\
  Universit{\'e} Clermont Auvergne, CNRS, ENSMSE, LIMOS\\ F-63000 Clermont-Ferrand, France \\
  \texttt{fabien.feschet@u-auvergne.fr} \\
}

\begin{document}
\maketitle

\begin{abstract}
In this paper, we propose an original manner to employ a tangential cover algorithm - minDSS - in order to geometrically reconstruct noisy digital contours. To do so, we exploit the representation of graphical objects by maximal primitives we have introduced in previous works. By calculating multi-scale and irregular isothetic representations of the contour, we obtained 1-D (one-dimensional) intervals, and achieved afterwards a decomposition into maximal line segments or circular arcs. By adapting minDSS to this sparse and irregular data of 1-D intervals supporting the maximal primitives, we are now able to reconstruct the input noisy objects into cyclic contours made of lines or arcs with a minimal number of primitives. In this work, we explain our novel complete pipeline, and present its experimental evaluation by considering both synthetic and real image data. We also show that this is a robust approach, with respect to selected references from state-of-the-art, and by considering a multi-scale noise evaluation process. 
\end{abstract}

% keywords can be removed

\section{Introduction}
\label{sec:intro}

The representation of graphical objects by geometrical primitives has been a long-term field of research, which appears at the birth of matrix screens and raster representation of images~\cite{Bresenham1965,Montanari1969}. This domain still stimulates the development of modern approaches with focus on different points like drawing vectorization~\cite{Bessmeltsev2019,Favreau2016}, deep learning based methods \cite{Egiazarian2020,Kim2018}, perception based clip-art vectorization~\cite{dominici2020polyfit}, from geometric analysis \cite{Kerautret2017}, or with real time constraint \cite{Xiong2017}. 

Image noise is a central problem in this complex task. To solve this issue, we can employ denoising algorithms to reduce this alteration as a pre-processing step~\cite{Lebrun2012}. Furthermore, segmentation methods~\cite{Wirjadi2007} can integrate smoothing terms, to control the behavior of active contours for instance~\cite{Kass1988}. Even if we can reach to smooth contours by such approaches, a lot of parameters must be finely tuned, depending on the input noise, and execution times can be very long. Modern deep learning approaches require large data sets to be tuned finely, for a specific application, and, sometimes, like for pixel art vectorization, the reference does not exist.

As a consequence, another strategy, that we adopt in this article, consists in calculating geometrical representations of noisy image objects, even if simple pre-processing and segmentation methods are employed first. 
Researches employing digital geometry notions and algorithms have mainly dealt with the reconstruction of graphical objects into line segments or circular arcs, originally by considering noise-free contours~\cite{Damaschke1995,Rosin1995}, later by processing noisy data~\cite{NGuyen2011a,NGuyen2010,Rodrigues2009a}. 

\begin{figure*}[htbp]
\centering
\includegraphics[width=.99\linewidth]{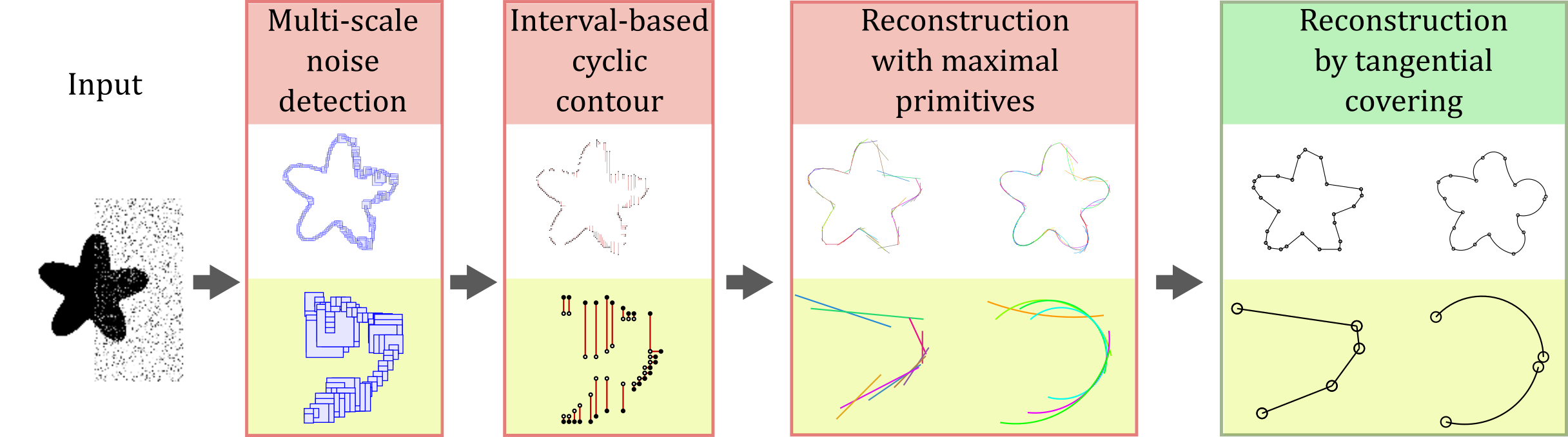}
\caption{Global work-flow of our reconstruction approach in relation to previous work~\cite{Vacavant2017} (in red) and new reconstruction, highlighted on the right (green).}
\label{fig:globalScheme}
\end{figure*}

Following this literature of digital geometrical-based approaches, in this article (an extended version of~\cite{Vacavant2021}), we propose a complete pipeline based on our previous contribution~\cite{Vacavant2017} in order to geometrically reconstruct noisy digital contours into line segments or circular arcs. This related work aimed at representing such contours with maximal primitives, by processing 1-D (one-dimensional) intervals obtained by a multi-scale and irregular approach, summarized in Fig.~\ref{fig:globalScheme}.

%%%
%%%
From this representation, we now propose to use the tangential cover algorithm minDSS introduced in~\cite{Feschet2005} to build a representative reconstruction composed of either lines or arcs. Originally, minDSS problem is the determination of the minimal length polygonalization (in term of input intervals) of a maximal primitives representation of a given regular eight-connected curve~\cite{Feschet2005}. In this article, we adapt this approach by using sparse irregular 1-D intervals as inputs. 

The article is organized as follows. Section~\ref{sec:maximal-primitives} relates synthetically on our previous contribution devoted to represent noisy digital contours into maximal primitives (lines or arcs)~\cite{Vacavant2017}. Then, Section~\ref{sec:tangential-cover} is dedicated to our novel approach that consists in adapting tangential cover algorithm for our specific contour representation, in order to reconstruct input objects into cyclic geometrical structures. Section~\ref{sec:results} presents the experimental results we have obtained for synthetic and real images, and a comparison with related works by considering robustness against contour noise. 
Improving the preliminary results~\cite{Vacavant2021}, this evaluation process employs a multi-scale noise approach according to the definition given in~\cite{Vacavant2018}.  
Finally, Section~\ref{sec:conclusion} concludes the paper with different axes of progress. 

\section{Reconstructions into maximal primitives}
\label{sec:maximal-primitives}
Since the reconstruction into maximal primitives is the main support
of the proposed representation, the main idea of the three
reconstruction steps are described in the following.

\subsection{Multi-scale noise detection}

\begin{figure}[htbp]
\centering
\subfloat[]{
\includegraphics[width=.2\linewidth]{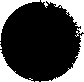}
\includegraphics[width=.2\linewidth]{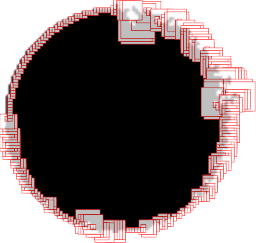}
}%%

\subfloat[]{
\includegraphics[width=.4\linewidth]{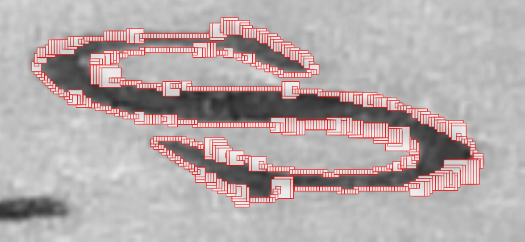}
\includegraphics[width=.4\linewidth]{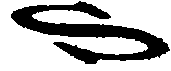}}
\caption{Examples of multi-scale noise detection upon a synthetic binary image (a) and a segmentation obtained from a real image (b).}
\label{fig:mboxes}
\end{figure}

We automatically detect the amount of noise present on a
digital structure by the algorithm detailed in~\cite{Kerautret2012}. 
From such a multi-scale analysis, the
proposed algorithm consists in constructing, for each contour point, a multi-scale profile defined by the segment length of all segments covering the point for larger and larger grid sizes. From each
profile, the noise level is determined by the first scale for which the slope of the profile is decreasing. This noise level can be represented as boxes and as exposed in Fig.~\ref{fig:mboxes}, a high noise in the contour will
lead to a large box, and \textit{vice-versa}. The algorithm can be
tested on-line from any digital contour given by a
netizen~\cite{Kerautret2014}. Also, Fig.~\ref{fig:mboxes} illustrates the possible noise altering image objects. In (a), a synthetic circle has been generated using various local noise levels for testing our approach, while in (b) the segmentation using a threshold leads to a noisy contour. In both cases, our pipeline can be applied since we consider a digital contour as input. Our first step leads  multi-scale boxes whose size depend on local noise amount. 
%%%

\subsection{Irregular isothetic cyclic representation}

The multi-scale contour representation calculated earlier is converted into irregular isothetic objects, defined in the general $d$-D case~\cite{Vacavant2018b} as: 
\begin{definition}[$d$-D $\mathbb I$-grid]
\label{def:igrid}
Let ${P}$ be a hyper-rectangular subset of $\mathbb{R}^d$.
A $d$-D $\mathbb I$-grid  denoted by  $G$ is a tiling of $P$ with rectangular hyper-parallelepipeds (or cells), which do not overlap, and  
whose faces of dimension $d-1$ are parallel to successive axes of the chosen space. 
A cell $R$ of $G$ is defined by a central point (position) $\mathbf{p}_R=(p^1_R,p^2_R,\dots,p^d_R)\in \mathbb{R}^d$ and a size along each axis $\mathbf{l}_R=(l^1_R,l^2_R,\dots,l^d_R)\in \mathbb{R^*_+}^d$. 
\end{definition}
In our context, we consider the 2-D case, as illustrated in Fig.~\ref{fig:ints-x-y}. Moreover, the input meaningful boxes are  converted into $k$-curves, formalized as:
\begin{definition}[$k$-curve]
\label{def:k-curve}
Let $A = (R_i)_{1\leq i\leq n}$ be a path from $R_1$ to $R_n$. $A$ is a $k$-curve iff each cell $R_i$ of $A$ has exactly two $k$-adjacent cells in $A$.
\end{definition}
In this definition, $k$-adjacency in 2-D is given by:
\begin{definition}[$k$-adjacency in 2-D]
\label{def:k-adj-2d}
Let $R_1$ and $R_2$ be two cells of a 2-D $\mathbb I$-grid $G$. $R_1$ and $R_2$ are $ve-$adjacent 
("vertex and edge" adjacent) if\ :
\begin{equation}
or \left\{
\begin{array}{c}
|x_{R_1}-x_{R_2}| = \frac{l^x_{R_1}+l^x_{R_2}}{2}\ and\ |y_{R_1}-y_{R_2}| \leq \frac{l^y_{R_1}+l^y_{R_2}}{2} \\[0.2cm]
|y_{R_1}-y_{R_2}| = \frac{l^y_{R_1}+l^y_{R_2}}{2}\ and\ |x_{R_1}-x_{R_2}| \leq \frac{l^x_{R_1}+l^x_{R_2}}{2}
\end{array}
\right.
\end{equation}
$R_1$ and $R_2$ are $e$-adjacent ("edge" adjacent) if we consider an exclusive "or" 
and strict inequalities in this definition.  
\end{definition}
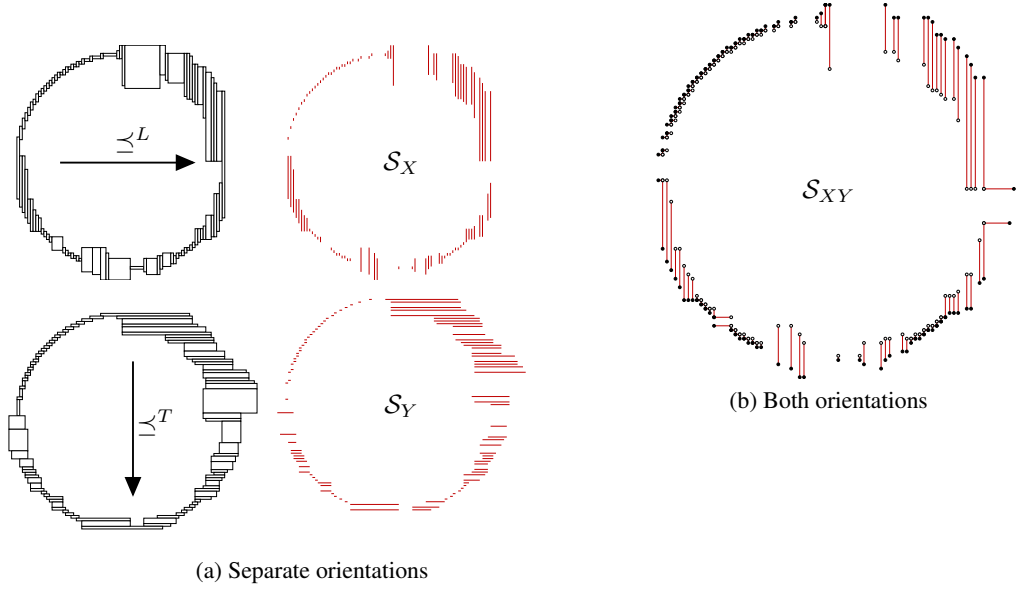
\begin{figure}
\centering
\setlength{\tabcolsep}{0.09cm}
\subfloat[Separate orientations]{
\begin{minipage}{.5\linewidth}
\begin{tabular}{cc}
\begin{tikzpicture}
    \draw (0, 0) node[inner sep=0] {\includegraphics[width=.4\linewidth]{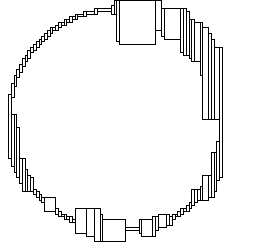}};
    \draw (-0.2, 0) node {\tikzstyle{mydiagramblock} = [draw, fill=yellow, text width=8em, minimum width=10em, text centered,
  minimum height=1.5em, rounded corners, drop shadow, text=black]
 \tikzstyle{line} = [draw, line width=0.25mm, color=black, -triangle 45]
 \tikzstyle{dash} = [dashed, line width=0.25mm, color=black, -triangle 45]
\begin{tikzpicture}[scale=0.8,transform shape,node distance=2.5cm, auto]
  \path node (p1) {};
  \path node (p2) [right of=p1] {};
  \path [line] (p1) -- (p2);
\end{tikzpicture}};
    \draw (0, 0.3) node {$\preceq^L$};
\end{tikzpicture} & 
\begin{tikzpicture}
    \draw (0, 0) node[inner sep=0] {\includegraphics[width=.4\linewidth]{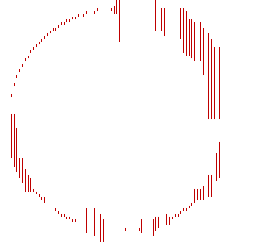}};
    \draw (0, 0) node {${\cal{S}}_X$};
\end{tikzpicture}
\\
\begin{tikzpicture}
    \draw (0, 0) node[inner sep=0] {\includegraphics[width=.4\linewidth]{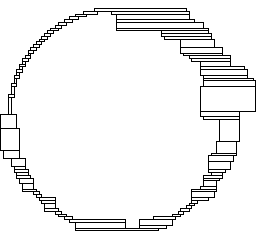}};
    \draw (0, 0) node {\rotatebox{-90}{\tikzstyle{mydiagramblock} = [draw, fill=yellow, text width=8em, minimum width=10em, text centered,
  minimum height=1.5em, rounded corners, drop shadow, text=black]
 \tikzstyle{line} = [draw, line width=0.25mm, color=black, -triangle 45]
 \tikzstyle{dash} = [dashed, line width=0.25mm, color=black, -triangle 45]
\begin{tikzpicture}[scale=0.8,transform shape,node distance=2.5cm, auto]
  \path node (p1) {};
  \path node (p2) [right of=p1] {};
  \path [line] (p1) -- (p2);
\end{tikzpicture}}};
    \draw (0.3, 0) node {$\preceq^T$};
\end{tikzpicture}
&
\raisebox{0.22cm}{
\begin{tikzpicture}
    \draw (0, 0) node[inner sep=0] {\includegraphics[width=.4\linewidth]{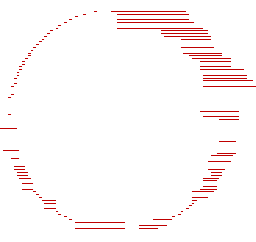}};
    \draw (0, 0) node {${\cal{S}}_Y$};
\end{tikzpicture}
}
\end{tabular}
\end{minipage}
}
\subfloat[Both orientations]{
%\begin{minipage}{.49\linewidth}
\raisebox{-1.cm}{
  %\multirow{2}{*}{
\begin{tikzpicture}
	\draw (0, 0) node[inner sep=0] {\includegraphics[width=.3\linewidth]{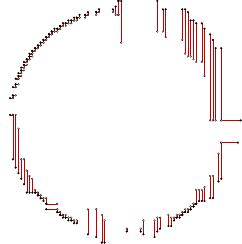}};
	\draw (0, 0) node {${\cal{S}}_{XY}$};
\end{tikzpicture}
%\end{minipage}
}
%}
}
\caption{Encoding of irregular isothetic cells into $k$-curves, following 2 order relations ($\preceq^L$ and $\preceq^T$), with interfaces stored inside the lists ${\cal{S}}_X$ and ${\cal{S}}_Y$. We then obtain a set of horizontal and vertical segments representing the input shape. In ${\cal{S}}_{XY}$, internal vertices are white, external ones are black.}
\label{fig:ints-x-y}
\end{figure}
Adjacency is defined as the spatial relationship between edges and vertices of two cells and may be compared to classic 4- and 8-adjacencies of regular square grids~\cite{Vacavant2018b}. In this article, we consider $k=ve$ wlog.

Furthermore, we have proposed to  calculate two $k$-curves, one for each axis, following the order relations  $\preceq^L$ ($X$ axis) and $\preceq^T$ ($Y$ axis)~\cite{Vacavant2018b} (see Fig.~\ref{fig:ints-x-y}). Then, both curves are combined to obtain 1-D $X$- and $Y$- aligned intervals stored inside the ${\cal S}_X$ and ${\cal S}_Y$ lists respectively. Finally, a single set ${\cal S}_{XY}$ is constructed by merging them, with a special care to respect the curvilinear abscissa of the input contour when ordering the intervals in this~list (see Fig.~\ref{fig:ints-x-y}~(b)). 

\subsection{Recognition of line segments and circular arcs}
\label{subsec:recog}
\begin{figure}[htbp]
\centering
\subfloat[Maximal segments]{
\includegraphics[width=.20\linewidth]{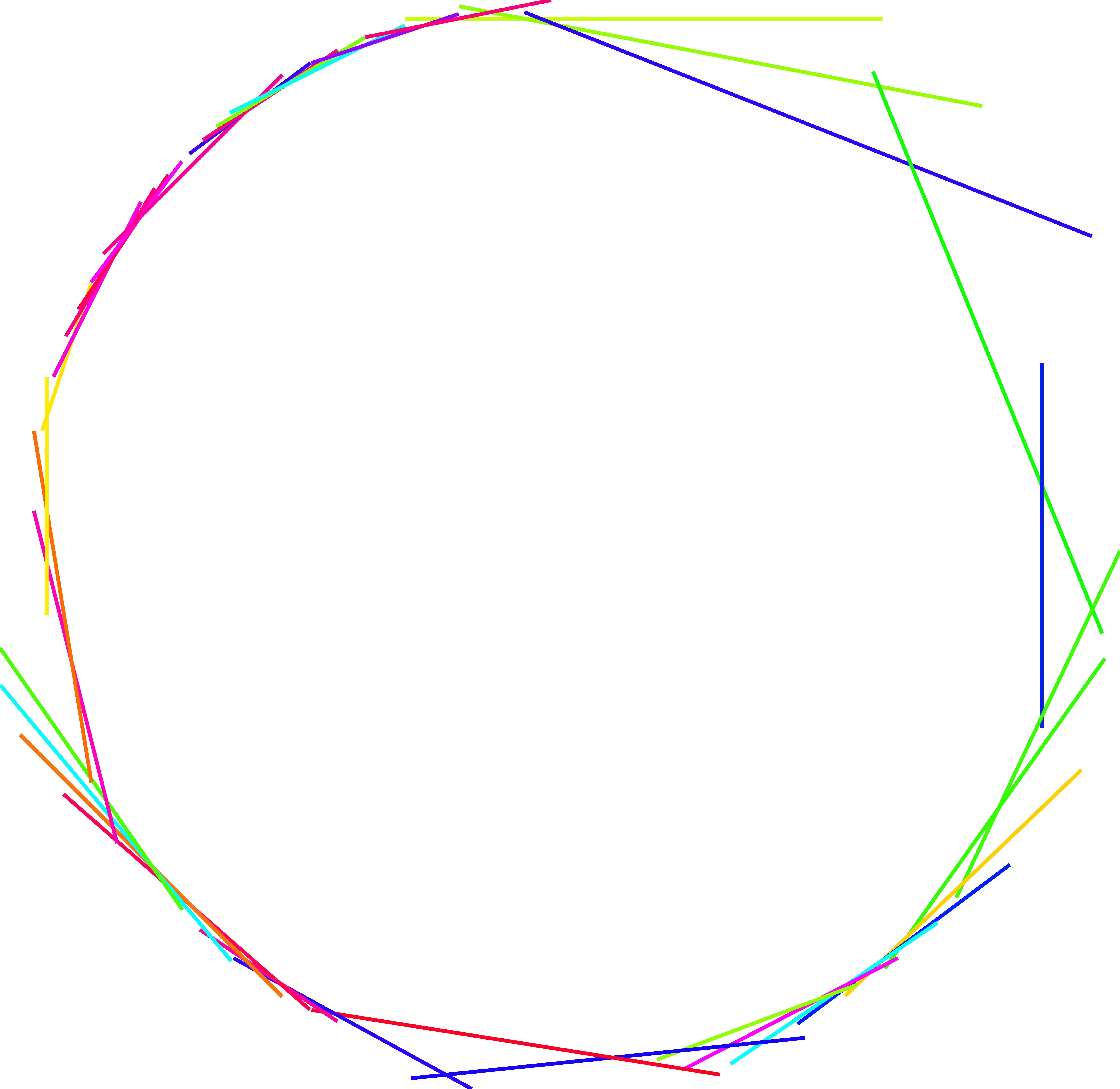}
\includegraphics[width=.45\linewidth]{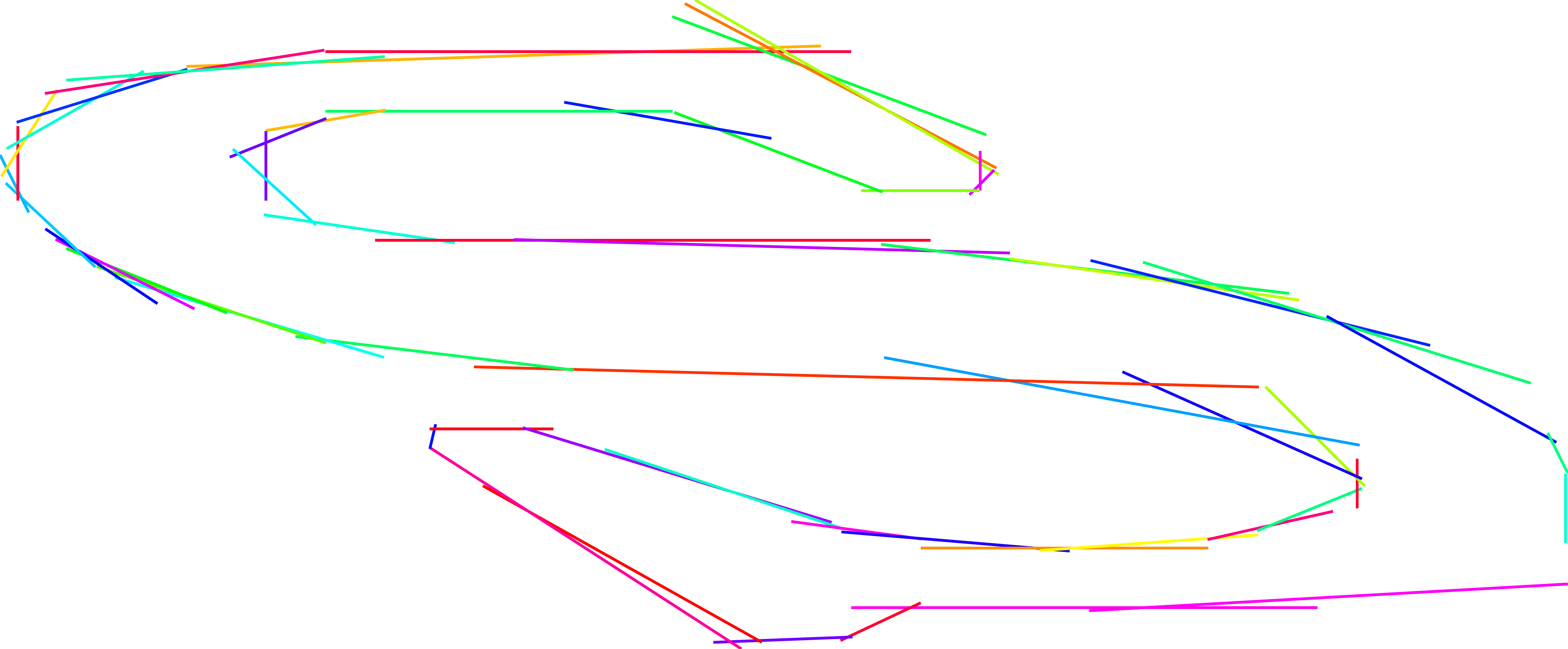}
}

\subfloat[Maximal arcs]{
\includegraphics[width=.2\linewidth]{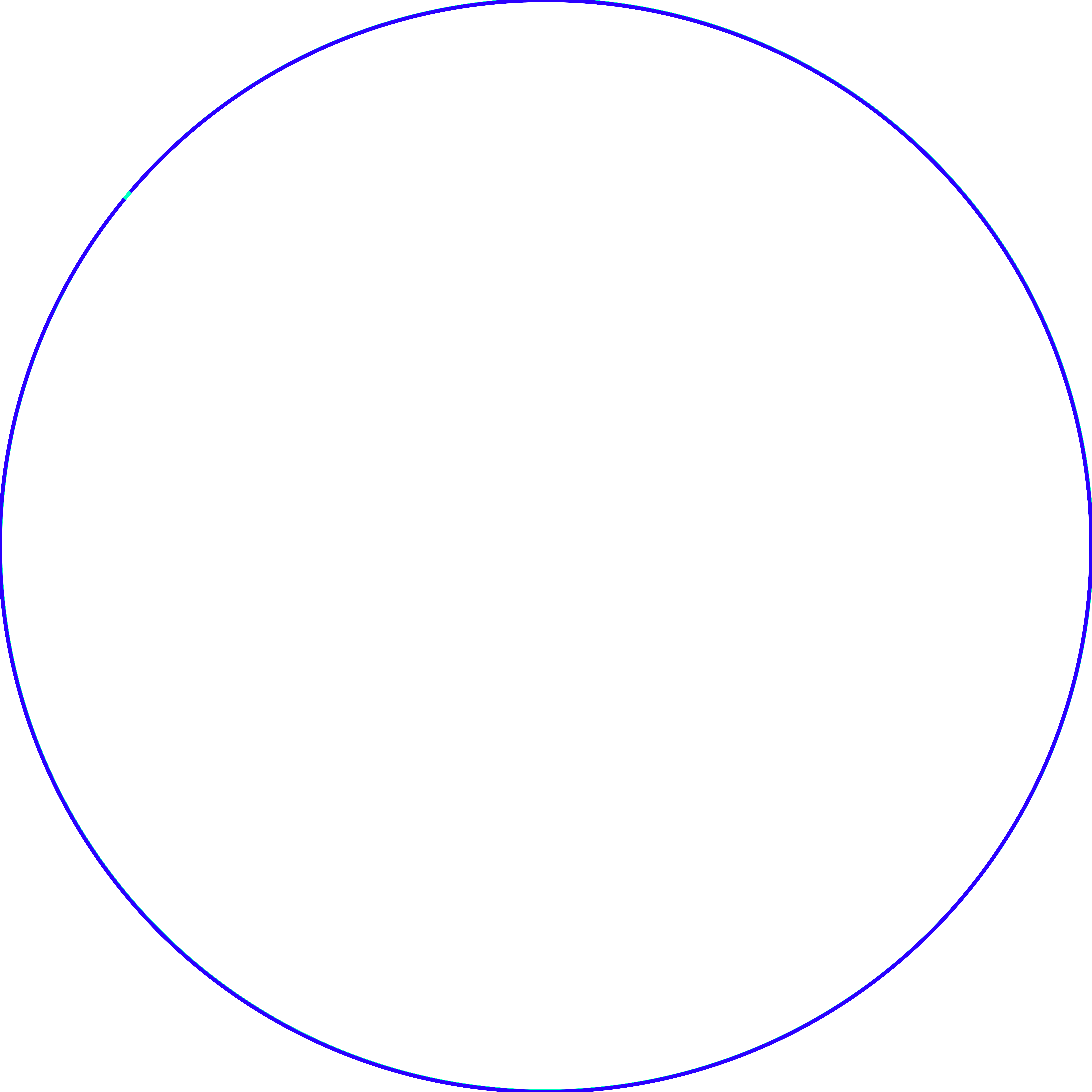}
\includegraphics[width=.45\linewidth]{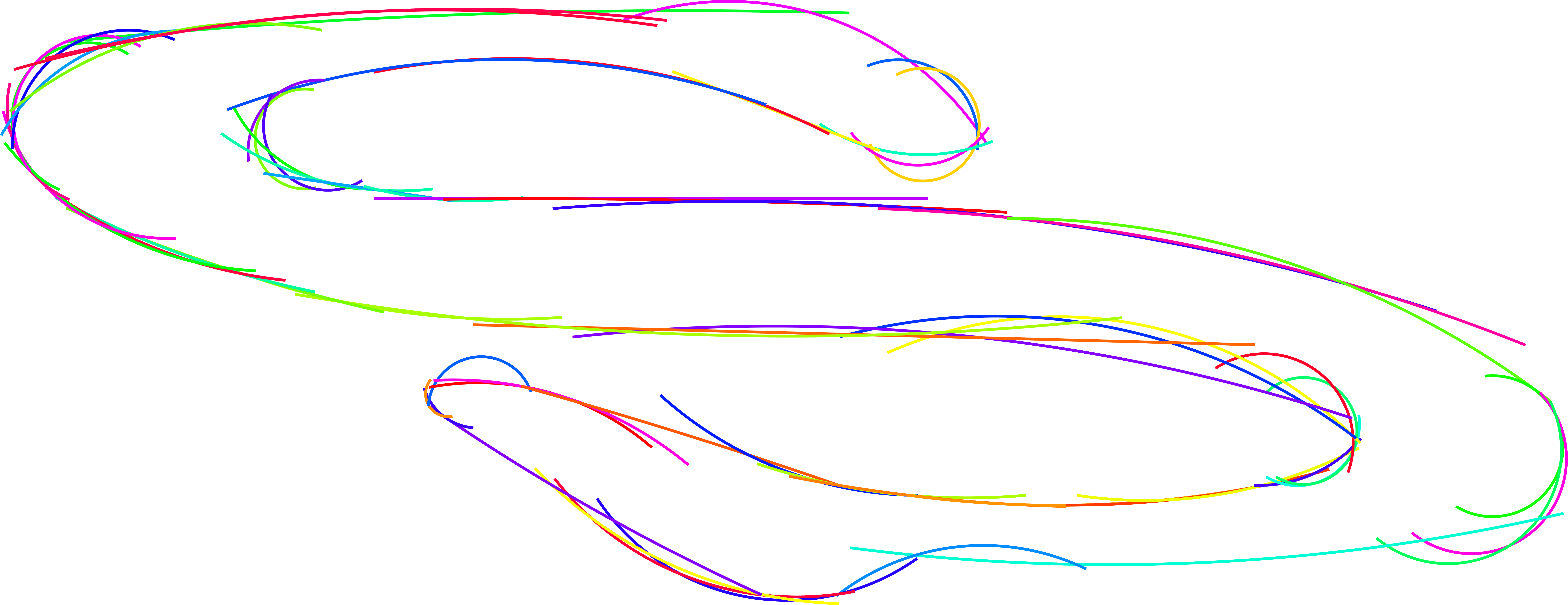}}
\caption{Reconstruction of digital shapes presented in Fig.~\ref{fig:mboxes} by maximal segments (a) and  arcs (b). Each primitive is presented with a random color.}
\label{fig:max-reconstruction}
\end{figure}

From this 1-D interval representation, we then obtain maximal segments or arcs by employing a GLP (Generalized Linear Programming) approach. In this step, we consider the internal and external points of the segments into two different sets $P^{\circ}$ and $P^{\bullet}$. This algorithm has been adapted from the works~\cite{Sharir1992,Welzl1991}, which aim to solve the problem of minimal enclosing circle. In our case, its purpose is to obtain primitives enclosing a set of points (\textit{e.g.} $P^{\circ}$)  but not the other  ($P^{\bullet}$). 
Fig.~\ref{fig:max-reconstruction} depicts the results obtained for two noisy digital contours, using maximal segments or arcs.  More details can be found in~\cite{Vacavant2017}.

%%%%%

%%%%%%

\section{Adapted tangential covering}
\label{sec:tangential-cover}
\subsection{Graph structure on ${\cal S}_{XY}$}
Using the results from the previous sections, at this step we have a collection of geometric primitives, either segments or circular arcs. This collection can be ordered according to the initial ordering of the constraints in ${\cal S}_{XY}$. The ordered set of primitives is induced by the constraints in ${\cal S}_{XY}$ such that its density is dependent on the density of the constraints. 
%Hence, there is no reason to have a uniform density. So the density is not linked to the underlying curve of the constraints. 
Since the density of 1-D intervals is not linked to the underlying curve, the constraints ${\cal S}_{XY}$ are non-uniformly distributed around the input contour. 
To provide a decomposition of the set of constraints into a sequence of consecutive geometric primitives, we can adapt the minDSS algorithm~\cite{Feschet2005}. More precisely, we will obtain a minimal length cycle of the set of primitives solely based on a graph structure induced by the solutions obtained in Section~\ref{subsec:recog}.

\begin{figure}
    \centering
    \includegraphics[width=.7\linewidth]{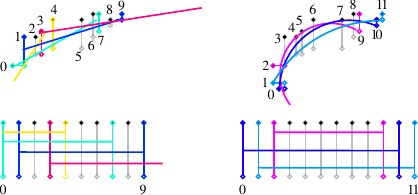}
    \caption{Ordering maximal primitives regarding their support upon the intervals, with segments (left) or arcs (right).}
    \label{fig:ordering}
\end{figure}

We first consider an arbitrary constraint in ${\cal S}_{XY}$ as the origin as well as an arbitrary orientation. An example of primitives computation is shown on Fig.~\ref{fig:ordering} for a part of a shape (from Fig.~\ref{fig:max-reconstruction}). To understand the graph structure embedded in ${\cal S}_{XY}$, we shall view each primitive as a consecutive set of constraints. For instance, the blue segment on the  Fig.\ref{fig:ordering}-left corresponds to the constraints 1 to 9. This precisely means that the subset of constraints numbered from 1 to 9 can be pierced by a segment but that this segment is not compatible with constraint indexed at 0 or greater than 9. Thus this primitive can be unambiguously represented by the interval of constraints $[1 ; 9]$. We now introduce a mapping from the constraints numbering onto the unit circle in $\mathbb{R}^2$ as follows: $k \mapsto \exp\lbrace \frac{2 i k\pi}{n} \rbrace$, where $n$ is the total number of constraints, that is $n=\mid {\cal S}_{XY} \mid$. As a consequence, each constraint is mapped onto a point on the unit circle and an interval of constraints is mapped onto an arc on the circle. We could thus view each arc as a node of a graph $G$ and there exists an edge between two nodes of the graph when their corresponding arcs overlapped. This graph $G$ is called a circular arc graph that will be processed to extract a minimal length cycle giving a series of consecutive primitives. 

%%%%
%%%%

\subsection{The minDSS algorithm}
\label{subsec:minDSS}
We now adapt the minDSS algorithm~\cite{Feschet2005} to solve the minimal length cycle problem in graph $G$. To understand the algorithm, we first recall that any path in graph $G$ corresponds to series of consecutive primitives due to the overlapping of arcs on $G$, associated to interval constraints on the shape. Since the set of constraints is ordered, the edges of $G$ are also ordered \textit{via} the mapping. Let us denote the node, \textit{i.e.} a geometric primitive reversing the mapping, of $G$ by $\{T_k\}_{k=0,n-1}$. We can hence call a node as a primitive to make intuition easier for the reader.
A cyclic ordering is used, which is the \textit{next} primitives following $T_k$ is $T_{(k+1)\mod{n}}$ and the \textit{previous} primitive is $T_{(k-1)\mod{n}}$. By iterations using \textit{next} and \textit{previous} operators, we can move along the set of primitives by using recursion as $\text{\textit{next}}^{(k)} = \text{\textit{next}}(\text{\textit{next}}^{(k-1)})$ and $\text{\textit{next}}^{(0)}=Id$ and obviously the same for \textit{previous}.

For a primitive $T_k$, we can define the \textit{forward} function $f(T_k)$ such that $f(T_k) = \text{\textit{next}}^{(j)}(T_k)$ with $T_k$ and $f(T_k)$ having overlapping arcs, hence connected in $G$, and $j$ maximal. It is a farthest forward move with overlapping of arcs. Since $T_k$ overlaps itself, this is well defined. By convention, we define $f^{-1}()$ similarly but using \textit{previous} instead of \textit{next}.

By construction $f^{-1}(f(T_k))$ is a primitive that overlaps $f(T_k)$. In a wide sense, it is a primitive before $T_k$ in the cyclic ordering, that is using \textit{previous} starting at $f(T_k)$. We can remark that the interval $[f^{-1}(f(T_k)) ; f(T_k)]$ is a separator of all cycles in the graph. Indeed, each cycle must contain a primitive in this interval. As a consequence, minDSS~\cite{Feschet2005} proposes to open the circular arcs in graph $G$ along the smallest of such separator to compute the shortest paths for each primitive in the chosen separator. To compute these shortest paths, we simply use iterations of the $f(.)$ function. Hence, the complexity of the shortest path is ${\cal O}(n)$ and if $\# F$ is the size of the smallest separator in $G$, the overall complexity is ${\cal O}(\#F \times n)$. To decide which is the shortest length cycle, we simply take the shortest one of the shortest path in the chosen separator. If the graph $G$ is proper, and this is the case for all reasonable notion of geometric primitives~\cite{Feschet2018}, the complexity reduces to ${\cal O}(n)$ since it is a decision problem and not a min computation.  

\subsection{Adaptation of minDSS}
\label{subsec:adaptminDSS}
In our approach, there is no underlying curve, that is, even if ${\cal S}_{XY}$ is a support set, it is not a curve. For instance, its density is not constant and is different from one as in the regular grid. Furthermore, this density can vary on different parts of the curve. Of course, minimizing the number of primitives, while having variable density, should be analyzed carefully. Indeed, minimizing the number of primitives led to different length steps in the underlying curve because length is directly linked with primitives' density. Hence, at the portions of the underlying curve with high density, using the primitives induces small movements while in the portions with low density changing from one primitive to the next one, can produce large movements in the curve. The fact that minDSS is actually a graph-based algorithm makes it independent of the embedding of the graph onto the underlying curve.  

\begin{figure}[htbp]
\centering
\subfloat[Line segments]{
\includegraphics[width=.9\linewidth]{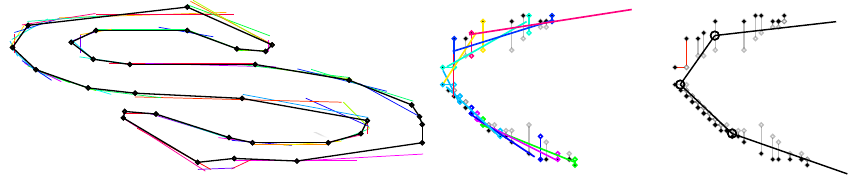}}

\subfloat[Circular arcs]{
\includegraphics[width=.9\linewidth]{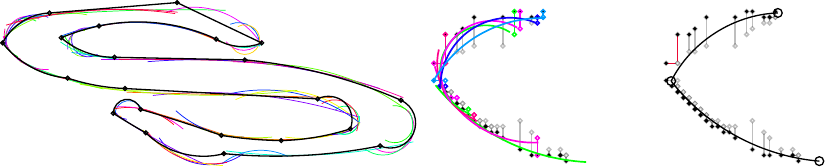}}
\caption{From the collection of maximal primitives obtained by GLP (random colors), minDSS leads to a set of potential reconstructions, for both kinds of primitives ; in this figure one of them is depicted in black. By considering the starting and ending intervals of these primitives (see bottom), minDSS produces cycles by minimizing the number of intervals. The intervals represented in red are examples of not intersected intervals.}
\label{fig:maxprims-tc}
\end{figure}

The output of the minDSS algorithm is a cycle in the set of intervals using the link between intervals made by the existence of a geometric primitive intersecting all of them. Hence, the output of minDSS is a list of intervals and not a set of primitives. As seen in Fig.~\ref{fig:maxprims-tc}, for segments as well as for arcs, from a list of intervals, there exists an infinite number of primitives corresponding to the given intervals. So, after the computation of the solution of minDSS, a geometric solution must be reconstructed by choosing primitives for all edges in the graph solution given by minDSS. There are several ways to do that and a fast and simple solution is obtained by using the middle points of the intervals (see Fig.~\ref{fig:maxprims-tc}). Of course, the consequence of such a choice is that some intervals might not be intersected by the primitives and this happens when the extreme points of an interval is a must use points for defining the original solution of compatibility between intervals (see for instance the red intervals of Fig.~\ref{fig:maxprims-tc}). Since there are several solutions for minDSS, each graph solution leads to a different geometric solution. 
%%%
%%%

\section{Experimental results}
\label{sec:results}

\subsection{Global overview of the method}

\begin{figure}[htbp]
\centering
%%\subfloat[]{
\begin{minipage}{.99\linewidth}
\centering
\subfloat[\emph{Fish} image]{
\includegraphics[width=.3\linewidth]{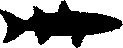}}
\subfloat[Meaningful scales]{
\includegraphics[width=.3\linewidth]{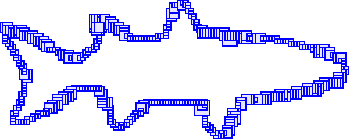}}
\subfloat[Intervals]{
\includegraphics[width=.3\linewidth]{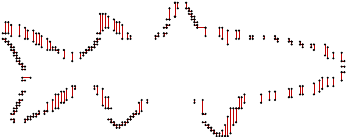}}
\\
\subfloat[Reconstructions with maximal segments and tangential cover]{
\includegraphics[width=.4\linewidth]{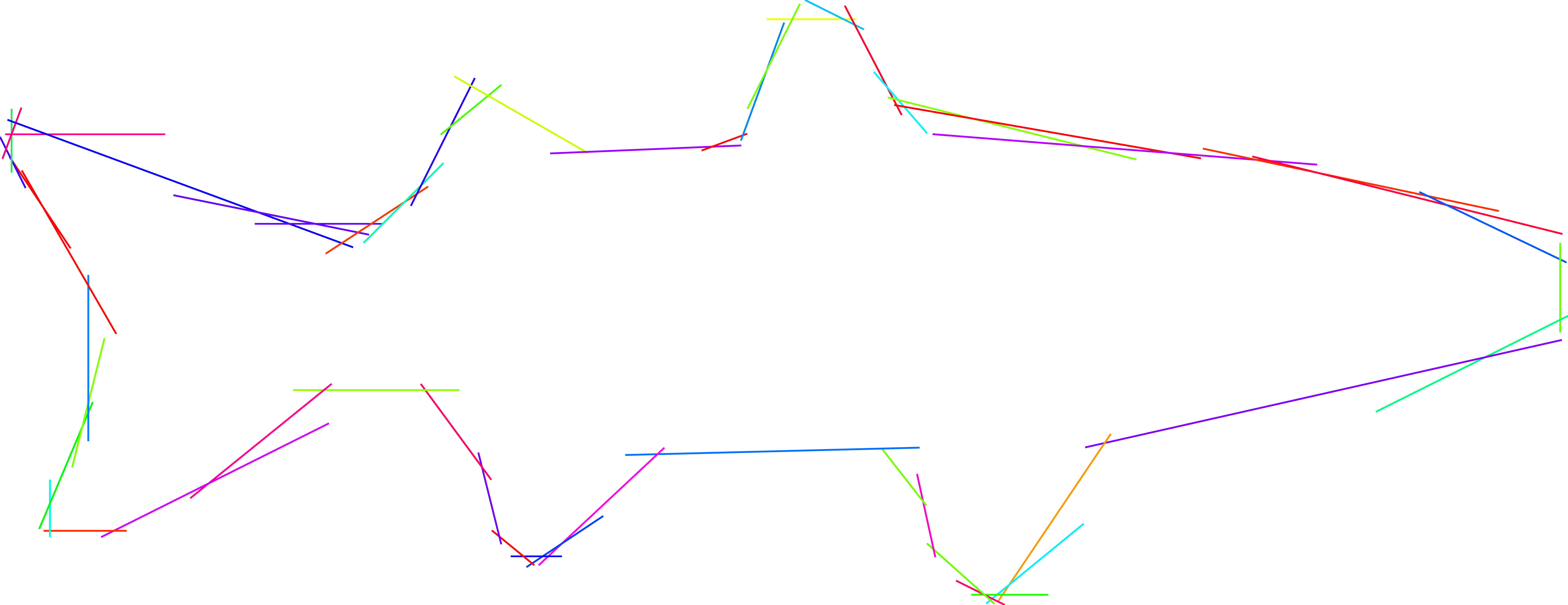}
\includegraphics[width=.4\linewidth]{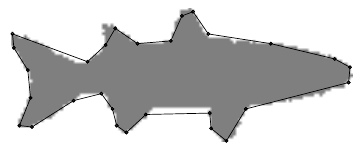}} \\
\subfloat[Reconstructions with maximal arcs and tangential cover]{
\includegraphics[width=.4\linewidth]{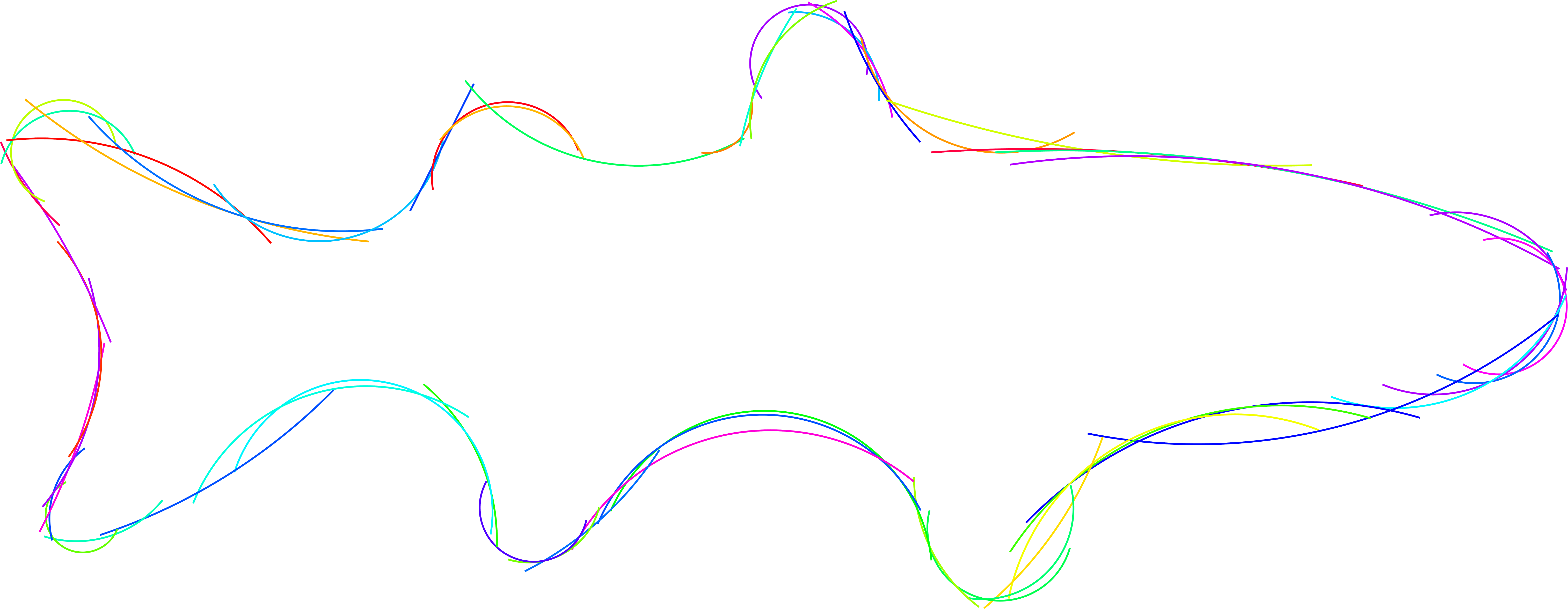} 
\includegraphics[width=.4\linewidth]{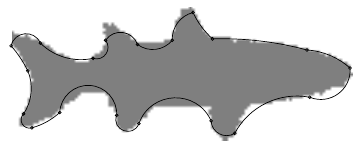}}
\vspace{.15cm}
\end{minipage}
%%}
\caption{Overview of our pipeline}
\label{fig:resultsa}
\end{figure}

We first propose to expose the different steps of our pipeline for a low resolution \emph{fish} binary image leading to a noisy shape (Fig.~\ref{fig:resultsa}-a), from the binary shape data set\footnote{\url{http://vision.lems.brown.edu/content/available-software-and-databases}}. We depict the meaningful scales (b), the set of intervals ${\cal S}_{XY}$ (c), the collections of maximal segments and arcs and the associated tangential cover outcomes (d,e). 
Thanks to our pipeline, we can obtain faithful reconstructions of such digital contours, with a low number of primitives. 

More precisely, when we consider segments, we can observe that straight parts of the shape are represented with line segments (\textit{e.g.} see the bottom part), since meaningful scales have a very fine resolution, and few intervals are thus required. More complex parts have been simplified faithfully (as triangular shapes for fish fins). 
When we select arcs instead, rounded parts of the shape are individually represented by circular primitives. Even if some regions are more smoothed (\textit{e.g.} see fish fins at top-left), these are relevant approximations for such a low resolution image (with no visual aberrant inverted arcs). Furthermore, it should be recalled that tangential cover computes one solution, and other options could be of better interest, depending on the final application.  

\subsection{Visual inspection of results with synthetic images}
\begin{figure}[htbp]
\centering
%%\subfloat[]{
\begin{minipage}{.99\linewidth}
\centering
\subfloat[Segment-based reconstruction of a polygonal shape]{
\includegraphics[width=.4\linewidth]{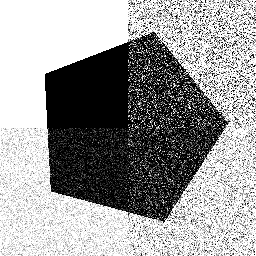} 
\includegraphics[width=.4\linewidth]{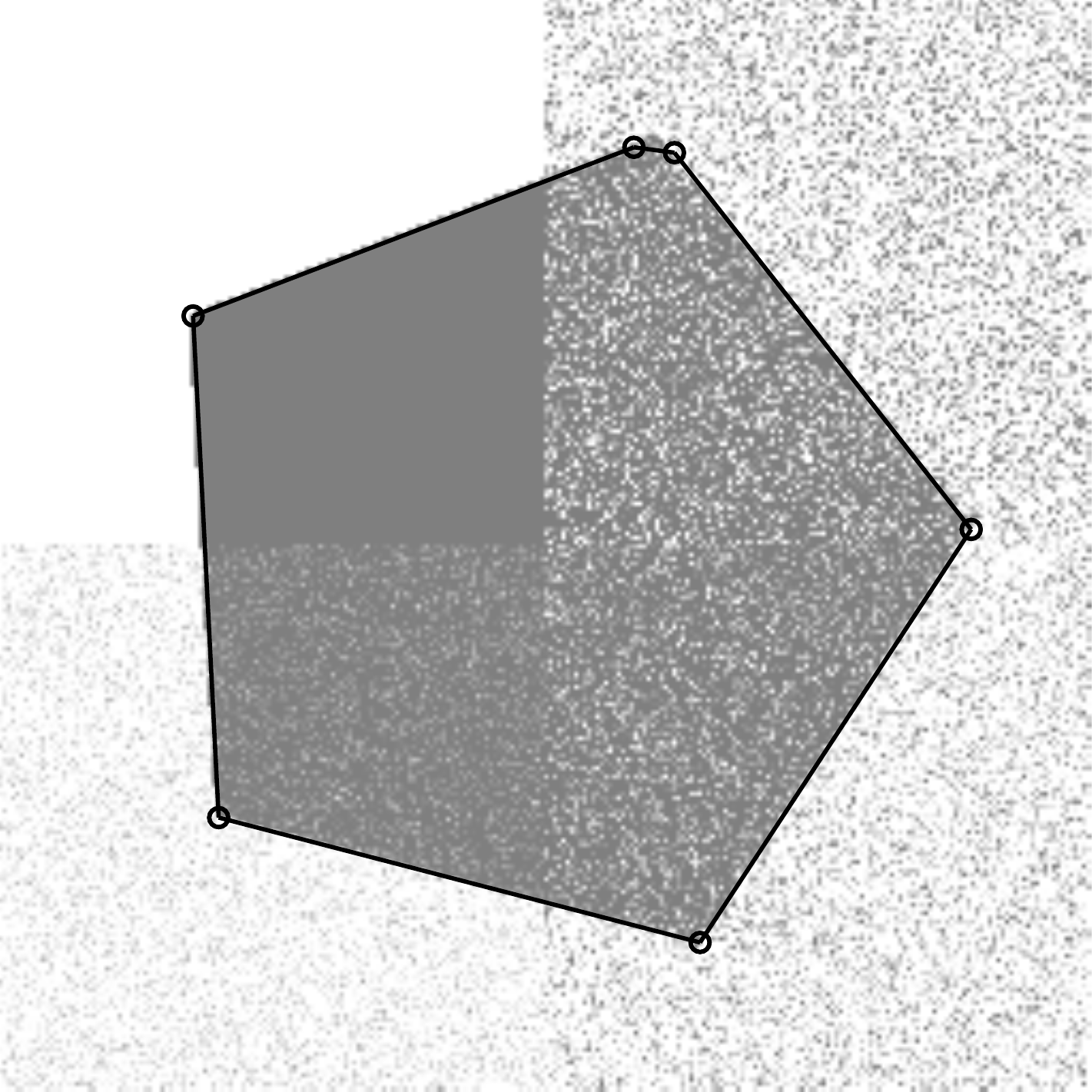}} \\
\subfloat[Arc-based reconstruction of a curved shape]{
\includegraphics[width=.27\linewidth]{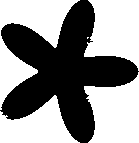} 
\includegraphics[width=.27\linewidth]{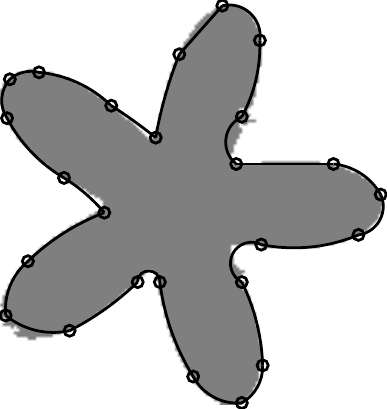}}
\end{minipage}
%%}
\caption{The reconstructions obtained for 2 synthetic images.}
\label{fig:resultsb}
\end{figure}

In Fig.~\ref{fig:resultsb}, two synthetic noisy objects, with polygonal and curved shapes, are respectively reconstructed with segments and arcs. 
In (a), a polygon image has been corrupted with different noises, to study the impact of variable perturbations into a single sample. We can notice that the final segment-based reconstruction is very stable and represents faithfully the underlying object. 
In (b), a curved shape is corrupted with local noises. Our algorithm can produce a relevant arc-based reconstruction, with a low number of primitives. As a whole, even in the presence of severe (possibly only local) perturbation, we can calculate reconstructions geometrically close to the underlying contours. 

\subsection{Visual inspection of results with a real image}

\begin{figure}[htbp]
\centering
%%\subfloat[]{
\begin{minipage}{.99\linewidth}
\centering
\subfloat[Meaningful scales]{
\includegraphics[width=.3\linewidth]{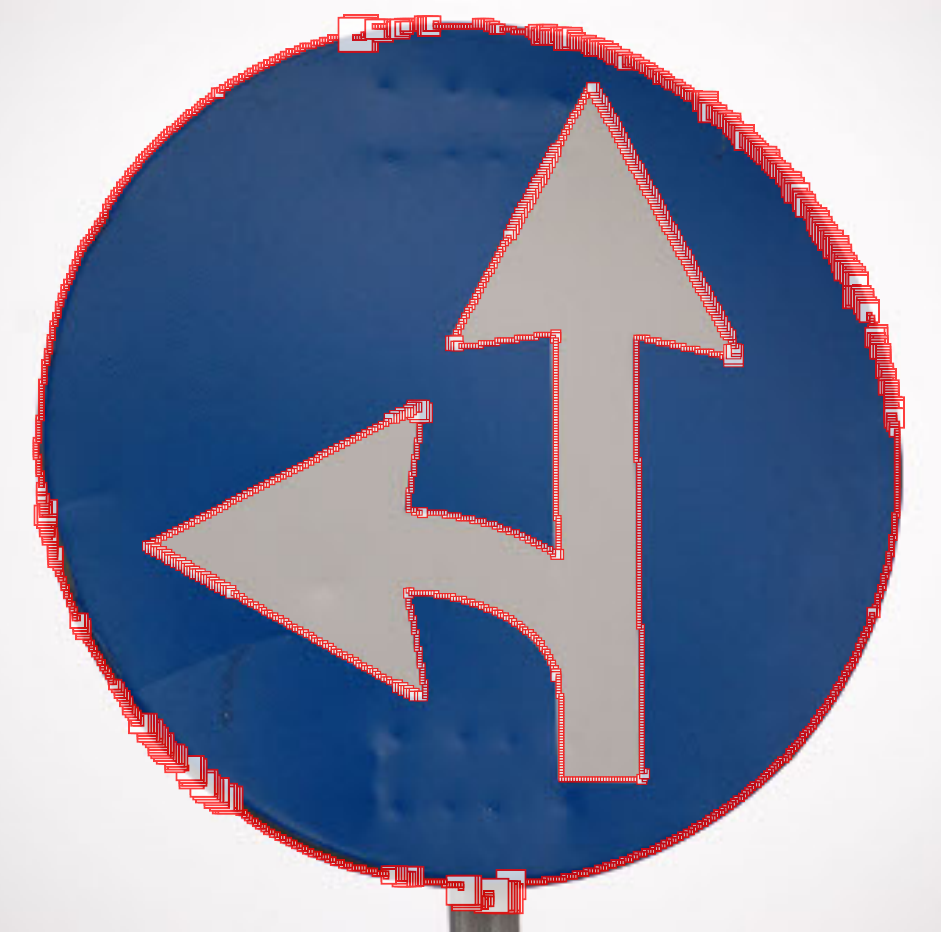}}
\subfloat[Maximal primitives]{
\includegraphics[width=.3\linewidth]{Figs/sign-maxsegs-maxarcs}} \\
\subfloat[Tangential cover]{
\includegraphics[width=.4\linewidth]{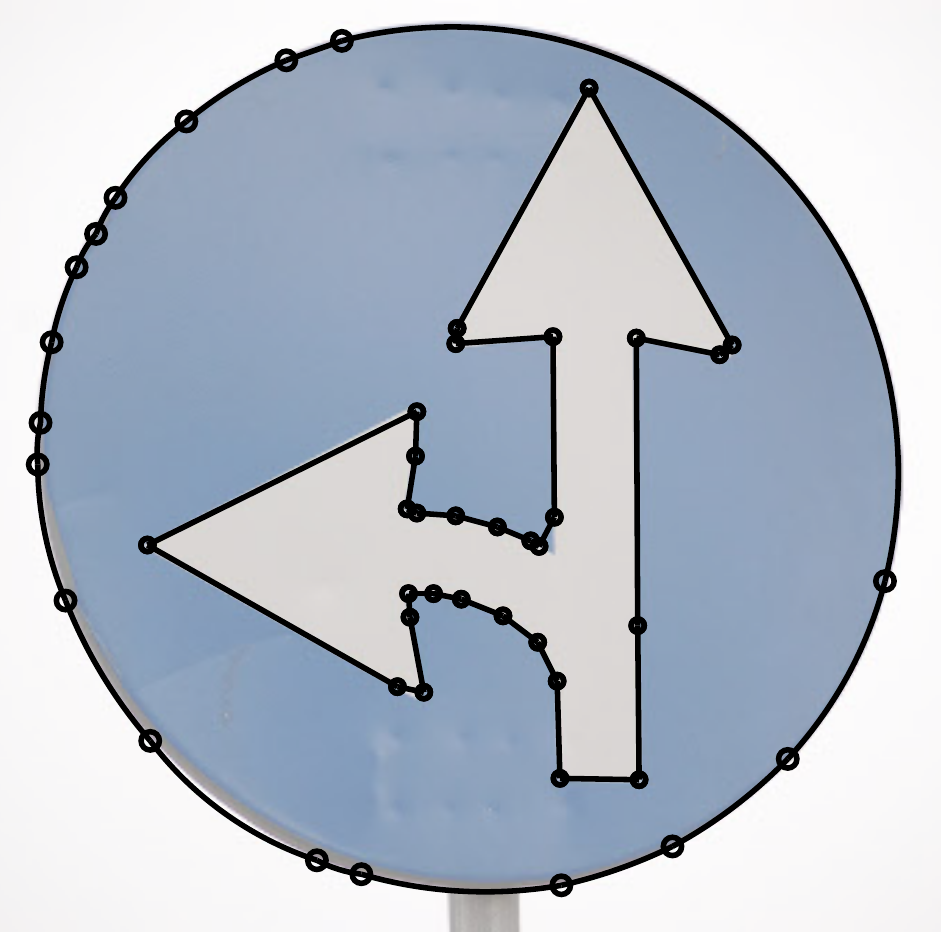}}
\end{minipage}
%%}
\caption{Visual evaluation of our pipeline with a large image}
\label{fig:resultsc}
\end{figure}

Finally, in Fig.~\ref{fig:resultsc}, we have selected a large image composed of two contours, processed independently by our approach. The external elliptical contour is reconstructed with arcs, while the internal one is represented with segments. 
We can see from the meaningful scales computed (a) that some parts of the contours are very finely represented (\emph{e.g.} see top-left part of the external part), which leads to small maximal primitives in (b) and in the tangential cover (c). On the contrary, noisy regions are reconstructed with longer primitives (\emph{e.g.} see straight lines of the internal part). 
Even if the input contours are long (in term of number of pixels)  tangential cover leads to reconstruction with few primitives.

\subsection{Comparative evaluation of robustness}
In this section, we propose to compare our contribution with related works (in the field of digital geometry), addressing graphical object reconstruction by line segments or circular arcs. To do so, we use the definition of robustness introduced in~\cite{Vacavant2018}, based on a multi-scale representation of image noise, which can be summarized as follows. 

Let $A$ be an algorithm dedicated to image processing, leading to an output $\mathbf{X}=\{x_i\}_{i=1,n}$ (in our case the geometrical reconstruction output). Let $N$ be an uncertainty specific to the target application of this algorithm, and  $\{\sigma_k\}_{k=1,m}$ the scales of $N$. In the following, we will consider a noise altering the contour. The different outputs of $A$ for every scale of $N$ is $\mathbb{X}=\{\mathbf{X_k}\}_{k=1,m}$. The ground truth is denoted by 
$\mathbb{Y}^0=\big\{\mathbf{Y_k^0}\big\}_{k=1,m}$. 
Let $Q({\mathbf{X_k}},\mathbf{Y_k^0})$ be a quality measure of $A$ for scale $k$ of $N$. This $Q$ function's parameters are the result of $A$ and the ground truth for a noise scale $k$. For our evaluation, we will consider the number of primitives obtained in the output. Obviously this measure is not necessary a guaranty of the resulting geometrical quality but it can give an indication on the stability of the reconstruction in particular on the specific test shapes used in the following. Our definition of robustness is expressed as follows: 
%%%%
\begin{definition}[$(\alpha,\sigma)$-robustness]
\label{def:robustness}
Algorithm $A$ is considered as robust if the difference between the output $\mathbb{X}$ and ground truth $\mathbb{Y}^0$ is bounded by a Lipschitz continuity of the $Q$ function:
\begin{equation}
\label{eqn:robustness-def}
\nonumber d_Y\left(Q(\mathbf{X_k},\mathbf{Y_k^0}),Q(\mathbf{X_{k+1}},\mathbf{Y_{k+1}^0})\right) \leq \alpha d_X(\sigma_{k+1}, \sigma_k),\ 1\leq k< m,
\end{equation}
where
\begin{eqnarray}
\nonumber
 d_Y\left(Q(\mathbf{X_k},\mathbf{Y_k^0}),Q(\mathbf{X_{k+1}},\mathbf{Y_{k+1}^0})\right) &=& Q(\mathbf{X_{k+1}},\mathbf{Y_{k+1}^0})-Q(\mathbf{X_k},\mathbf{Y_k^0}), \\
\nonumber  d_X(\sigma_{k+1}, \sigma_k) &=& |\sigma_{k+1} - \sigma_k|.
\end{eqnarray}
We calculate the robustness measure $(\alpha,\sigma)$ of $A$ as the $\alpha$ value obtained and the scale $\sigma=\sigma_k$ where this value is reached.  
\end{definition} 
%%%%
In other words, $\alpha$ measures the worst difference in quality through the scales of uncertainty $\{\sigma_{k}\}$, and $\sigma$ keeps the noise scale leading to this value. The most robust algorithm should have a low $\alpha$ value, and a very high $\sigma$ value.

\begin{figure}[htbp]
\centering
\subfloat[$\theta_2=\pi/8,\ k=5$]{\includegraphics[width=.3\linewidth]{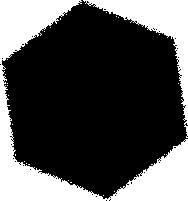}}\quad
\subfloat[$\theta_5=\pi/2,\ k=10$]{\includegraphics[width=.3\linewidth]{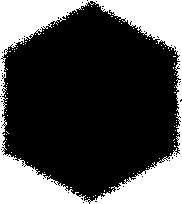}}

\subfloat[$\theta_1=0,\ k=1$]{\raisebox{.5cm}{\includegraphics[width=.37\linewidth]{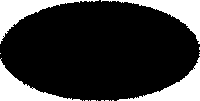}}}\quad
\subfloat[$\theta_3=\pi/4,\ k=7$]{\includegraphics[width=.3\linewidth]{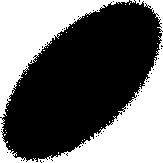}}
\label{fig:rob-shapes}
\caption{Samples of shapes employed in our tests of robustness.}
\end{figure}
\begin{figure}[htbp]
\centering
\includegraphics[width=.8\linewidth]{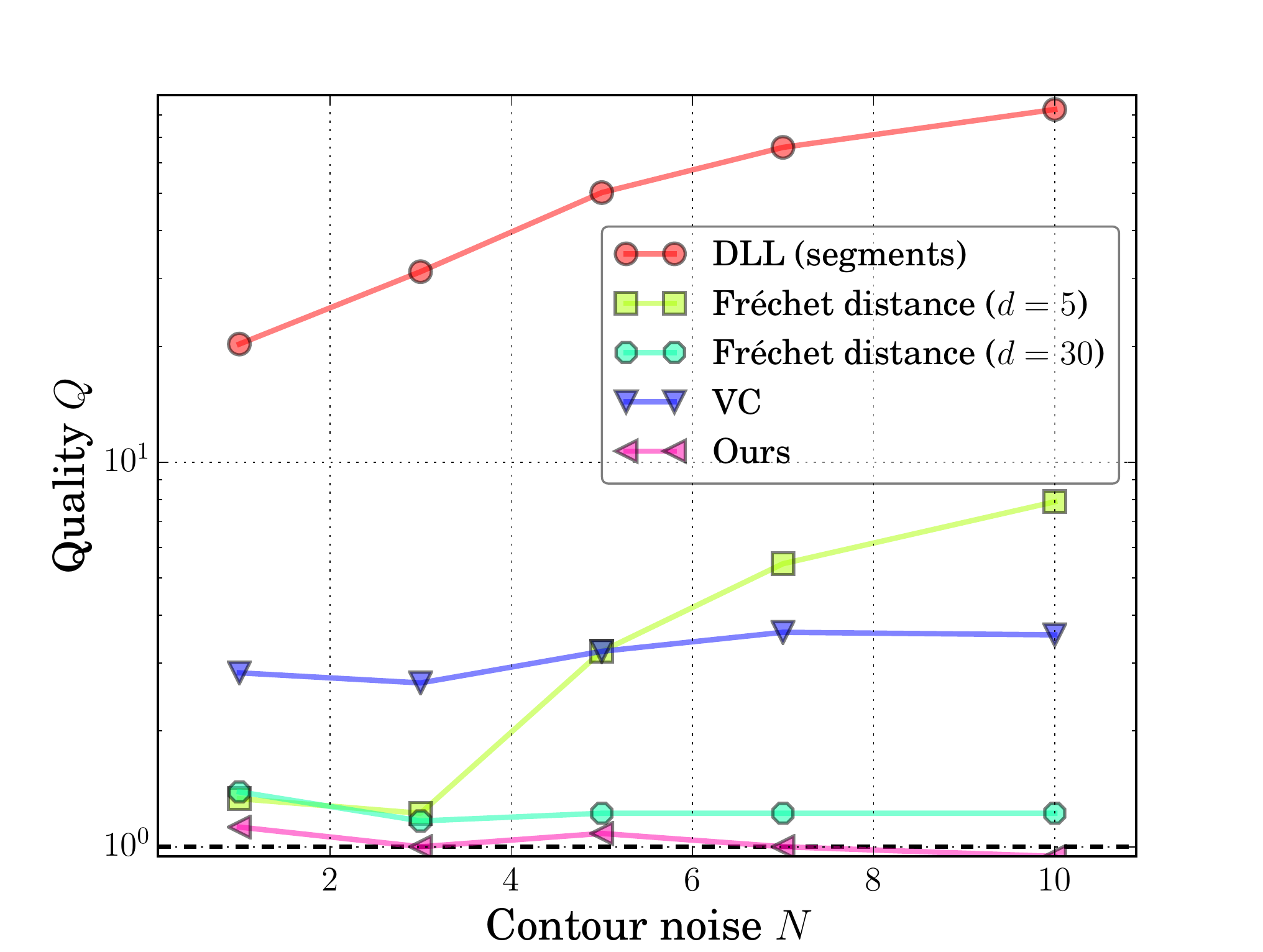}

\small
\begin{tabular}{l|c|r}
\hline
\bf Name & \bf Ref. & \bf $(\alpha,\sigma)$ \\
\hline
\hline
DLL (segments) & \cite{Provot2014} & (9.472,3.0) \\
Fr\'echet distance ($d=5$) & \cite{Sivignon2014} & (1.111,5.0)\\
Visual curvature & \cite{Liu2008} & (0.278,3.0) \\
Fr\'echet distance ($d=30)$  & \cite{Sivignon2014} & (0.111,1.0) \\
Ours  & -- & (0.063,1.0) \\
\hline
\end{tabular}
\caption{Graphical (top) and numerical (bottom) evaluations of robustness of our algorithm with line segments. }
\label{fig:rob-lines}
\end{figure}

We first propose to study the efficiency of our algorithm for calculating reconstruction with line segments. We use a synthetic polygonal shape, increasingly altered with a Kanungo-based contour noise~\cite{Kanungo2000} (see Fig.~\ref{fig:rob-shapes}-a,b). With our notations, this means that $N$ is modeled with scales $\{\sigma_k\}_{k=1,5}=\{1,3,5,7,10\}$. For each noise scale, we also consider a set of rotated shapes under angles $\mathbf{\Theta}=\{\theta_t\}_{t=1,5}=\{0,\pi/8,\pi/4,3\pi/8,\pi/2\}$. The $Q$ quality function compares the number of primitives obtained for a given algorithm $A$ for noise scale $k$ and angle $\theta_t$, denoted by $p^t_k$, with the objective number of primitives, $\mathbf{Y}^0=p^*$, by:
\begin{equation}
    \label{eqn:q-function}
    Q(\mathbf{X_k},\mathbf{Y_k^0})=\frac{1}{5p^*}\sum_{t=1,5}p^t_k,\ 1\leq k< m.
\end{equation}
In this first experiment, $p^*=6$ since we consider an hexagonal shape. In Fig.~\ref{fig:rob-lines}, we show the graphical and numerical evaluations of robustness we have obtained by comparing with the methods: Visual curvature (VC)  proposed by~\cite{Liu2008} (with parameter $s$ tested with different values but without a significant impact upon results); a Fr\'echet distance-based approach~\cite{Sivignon2014} (with parameter $d$ at two values, 5 and 30); the Digital Level Layers (DLL) with straight lines~\cite{Provot2014}. 

\begin{figure}[htbp]
\centering
\includegraphics[width=.8\linewidth]{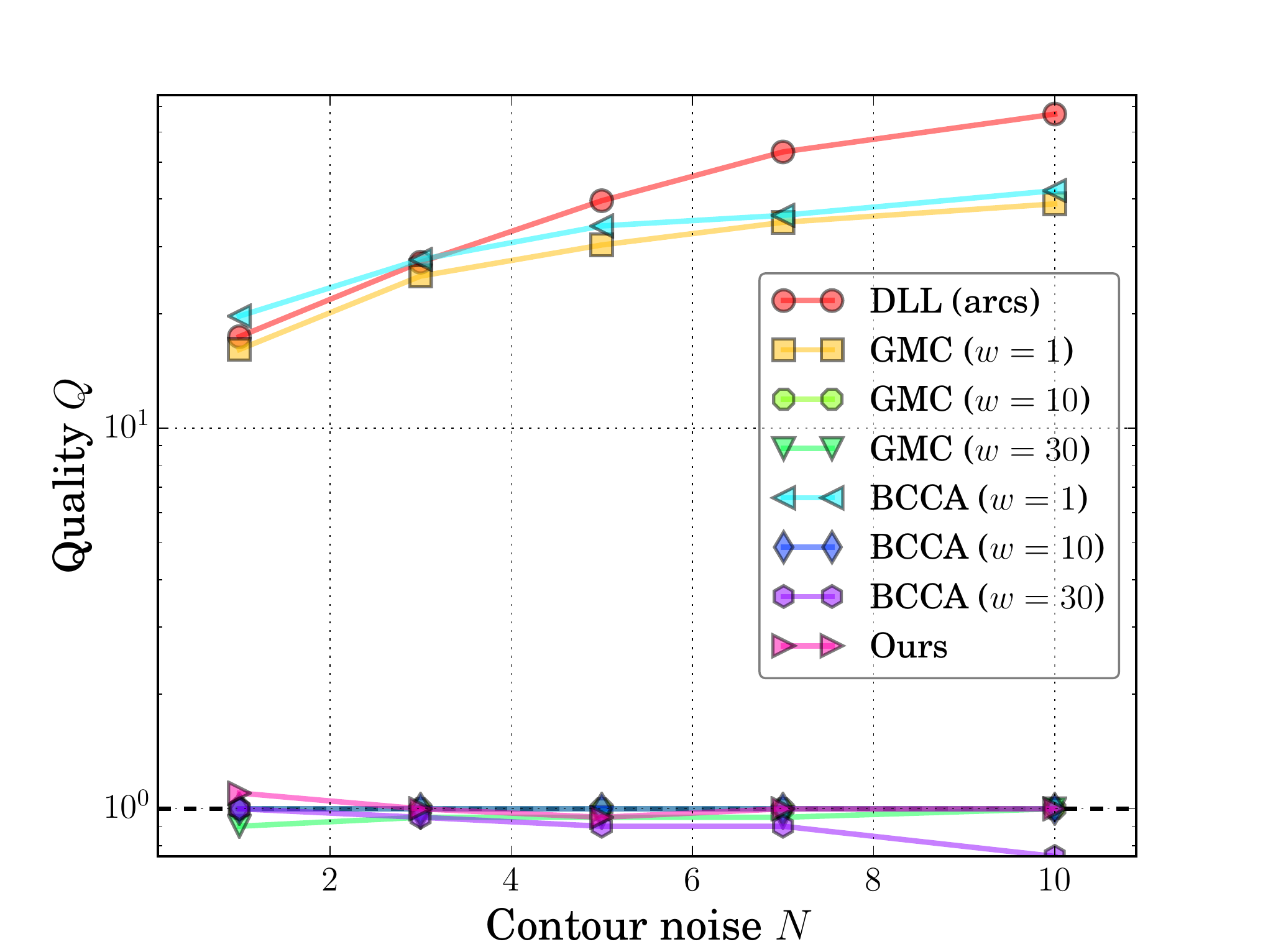}

\small
\begin{tabular}{l|c|r}
\hline
\bf Name & \bf Ref. & \bf $(\alpha,\sigma)$ \\
\hline
\hline
DLL (arcs)  & \cite{Provot2014} & (6.8,5.0) \\
GMC ($w=1$) & \cite{Kerautret2009} & (4.5,1.0) \\
BCCA ($w=1$) & \cite{Malgouyres2008} & (4.025,1.0) \\
Ours & -- & (0.05,1.0) \\
BCCA ($w=30$) & \cite{Malgouyres2008} & (0.05,7.0) \\
GMC ($w=30$) & \cite{Kerautret2009} & (0.025,1.0) \\
BCCA ($w=10$) & \cite{Malgouyres2008} & (0.0,1.0) \\
GMC ($w=10$) & \cite{Kerautret2009} & (0.0,1.0) \\
\hline
\end{tabular}
\caption{Graphical (top) and numerical (bottom) evaluations of robustness of our algorithm with circular arcs. }
\label{fig:rob-arcs}
\end{figure}

To evaluate our method with circular arcs, we have chosen an ellipse, which should be reconstructed as 4 primitives (\textit{i.e.} $p^*=4$). Thanks to the same methodology (rotations of the shape $\mathbf\Theta$, set of noise scales $N$), we have compared our approach with DLL (with circles)~\cite{Provot2014}, GMC~\cite{Kerautret2009} based~\cite{kerautret_curvature_2008} (with parameter $w$ at 1, 10 and 30), BCCA~\cite{Malgouyres2008} based~\cite{kerautret_curvature_2008} (with the same parameters). Fig.~\ref{fig:rob-arcs} presents the results we obtained for all the tested methods. 
%%%

Both experiments show that our contribution is a robust approach, for segments and arcs. Its $(\alpha,\sigma)$ values are close (even superior for segments) to the most accurate techniques (Fr\'echet distance-based and BCCA/GMC) but without any parameter to be given \textit{a priori}. Graphically, we can notice that our method stands close to the objective value (1 in dotted lines), while the less robust algorithms have a number of primitives increasing dramatically with the augmentation of noise.

\section{Conclusion and future works}
\label{sec:conclusion}
 
We have proposed a novel framework to reconstruct possibly noisy digital contours by line segments or circular arcs, by combining our previous work devoted to reconstruct maximal primitives~\cite{Vacavant2017} and the tangential cover algorithm minDSS~\cite{Feschet2005}. By means of the experiments we have exposed, that this method achieves faithful reconstructions according to the underlying object contour. Moreover, by employing minDSS, we ensure that the number of primitives is~low.

Our first research lead is to compare our method with other classic and modern reconstruction approaches, possibly dedicated to vectorization, by considering various quality measures as we did in~\cite{Vacavant2013} (Hausdorff distance, error of tangents' orientation, Euclidean distance between points). We would like to show that our pipeline permits to obtain robust reconstructions (according to the input contour noise), with respect to the state-of-the-art. 

Another important concern is to calculate a single reconstruction combining segments and arcs. To do so, we would like to propose a combinatorial optimization process as developed in~\cite{Faure2009}. Such scheme would select the best primitive to be integrated in the final solution, depending on a given predicate, \textit{e.g.} the geometrical length of the primitive. An alternative option is to first compute a set of maximal primitives containing both types, by considering the straight and curved part of the input noisy shape, as we employed already in~\cite{Vacavant2013}.  

Finally, we plan to compare the future reconstructions with other approaches able to choose the primitive automatically like the method proposed by Nguyen and Debled \cite{nguyen_decomposition_2011} and to explore other applications like the one related to 3-D printing \cite{valdivieso2018polyline}. 

\bibliographystyle{unsrt}  
\bibliography{refs}

\end{document}